# Service-Oriented Architecture for Weaponry and Battle Command and Control Systems in Warfighting

Youssef Bassil

LACSC – Lebanese Association for Computational Sciences
Registered under No. 957, 2011, Beirut, Lebanon
youssef.bassil@lacsc.org

## ABSTRACT

Military is one of many industries that is more computer-dependent than ever before, from soldiers with computerized weapons, and tactical wireless devices, to commanders with advanced battle management, command and control systems. Fundamentally, command and control is the process of planning, monitoring, and commanding military personnel, weaponry equipment, and combating vehicles to execute military missions. In fact, command and control systems are revolutionizing as war fighting is changing into cyber, technology, information, and unmanned warfare. As a result, a new design model that supports scalability, reusability, maintainability, survivability, and interoperability is needed to allow commanders, hundreds of miles away from the battlefield, to plan, monitor, evaluate, and control the war events in a dynamic, robust, agile, and reliable manner. This paper proposes a service-oriented architecture for weaponry and battle command and control systems, made out of loosely-coupled and distributed web services. The proposed architecture consists of three elementary tiers: the client tier that corresponds to any computing military equipment; the server tier that corresponds to the web services that deliver the basic functionalities for the client tier; and the middleware tier that corresponds to an enterprise service bus that promotes interoperability between all the interconnected entities. A command and control system was simulated and experimented and it successfully exhibited the desired features of SOA. Future research can improve upon the proposed architecture so much so that it supports encryption for securing the exchange of data between the various communicating entities of the system.

**Keywords:** *Service-Oriented Architecture, Computational Military, Command & Control, Web Service*

## 1. INTRODUCTION

Computing technologies are becoming more pervasive day after day, offering new potentials for automating tasks in many challenging applications. Military is one of these applications that is evolving at a quickening pace. It includes the use of computers to help support in decision making, tactic forecasting, ballistic trajectory calculations, direction transfer, navigation control, data and communication encryption, and command and control [1]. In essence, command and control, abbreviated as C2, is a battle management process by which military personnel, weaponry devices, fighting vehicles, military equipment, and communication and navigation facilities are commanded to achieve military aims and objectives [2]. In effect, many of military devices and weaponry equipment are highly computing intensive systems that use complex embedded software and algorithms to handle computational and data-intensive tasks. These days, it is no longer practical to develop military ad-hoc systems that require a single person to wisely craft the entire software for the military hardware equipment. In addition, it is no more feasible to encapsulate all software components within the actual equipment. Instead, a component-based model or service-oriented architecture is often followed in which software is developed as a set of services by multiple persons working on a large code base in a distributed team [3].

Inherently, a service is a software component that contains a collection of related software functionalities reusable for different purposes [4]. It delivers such operations as data storage, data processing, mathematical and scientific computations, and networking. It is governed by a producer-consumer model in which a service is delivered by a service provider known as the producer which owns the facilities for hosting, running, and maintaining the service, and the client known as the consumer which connects and uses service functionalities via remote method invocation mechanism. Predominantly, services are implemented as Web Services (WS) which are defined by the W3C as "software systems designed to support interoperable machine-to-machine interaction over a network" [5].

This paper proposes a service-oriented architecture for weaponry and battle command and control systems in war fighting based on heterogeneous multi-platform service components. The proposed architecture is composed of three basic tiers: The first tier is the client represented by the military hardware equipment. The second tier is the server which hosts and runs the different service components that provide the advanced functionalities necessary for the operation of the client equipment. The third tier is the middleware represented by an Enterprise Service Bus (ESB) which offers a standard interface and a data-path for both the client and server tiers to interact, send requests, and receive responses from each other.

Being decentralized and decoupled from the military equipment hardware core, the proposed service-oriented architecture has six benefits [6][7][8]: Integrate-ability which allows the seamless integration of new software components in a less significant effort, time, and budget; reusability which is given by the nature of SOA "build once, use many times" that allows multiple military equipment, possibly located in different sites, to use and share the same set of services simultaneously and with high availability; scalability which is given by the ability to add, update, and delete military equipment's functionalities remotely with no or minimal service interruption and while the system is online; maintainability which is given by that a failure in a service would only require replacing the faulty service and not the entire battle command system; survivability which is given by that service components in SOA are decentralized and thereby they can be replicated across military data centers allowing military systems to withstand a hit and remain mission-capable during the war time; and interoperability which is given by the Enterprise Service Bus middleware which provides a standardized and a unified platform for the various interconnected entities, possibly incompatible, to send and receive data among each other.

## 2. BATTLE COMMAND & CONTROL

Fundamentally, battle command (BC) also known as command and control (C2), is the science and practice of commanding, controlling, describing, directing, and leading military forces and combatting machineries during war fighting [9]. It involves military decisions and processes that are initiated by commanders through computing and communication facilities and executed by soldiers located in remote areas in the war zone with the purpose of accomplishing a desired military objective or mission. Generally, battle command is managed through a command and control center or command post often located in a secure building operated by governmental or military agencies. In modern warfare, C2 is extended to support in addition to command and control, other features and functionalities such as reconnaissance, intelligence, surveillance, communications, computers, information systems, and target acquisition. These improved versions of C2 are denoted by a number of abbreviations in the format $C^{(x)}$ followed by supplementary letters indicating the supported features. For instance the $C^5$ISTAR system stands for command, control, communications, computers, combat systems, intelligence, surveillance, target acquisition, and reconnaissance [10]. By definition, command is the use of authority to achieve a particular objective. Control is the process of guiding, validating, and refining actions based on the objective to be accomplished. Communication is the process of conveying the command and control to the destination unit. Computer is the use of computing facilities to perform data processing to support commanders' decision-making. Combat systems designate the process of operating and managing military equipment, devices, and machineries in the battlefield. Intelligence is the process of collecting, analyzing, and assessing facts, data, and information. Surveillance is the process of monitoring the behavior and activities of certain subjects. Target acquisition is the process of detecting, identifying, and locating military targets. Reconnaissance is the process of exploring enemy forces to gain information about their environments and assets [11].

Practically, communication between the battle command centers and the fighting units is done through communications satellites or COMSATs which are artificial satellite positioned in space in geostationary orbits, low earth orbits, and other elliptical orbits for the purpose of conveyance of information by armed forces in a reliable, fast, secure, and jam-resistant manner.

Traditional battle control architectures are platform-centric [12], in that, military equipment supporting digital computation such as artillery controllers, missiles, warheads, warships, submarines, combat vehicles, aircrafts, traffic control radars, surveillance sensors, and GPS systems incorporate their software into their core hardware. In this type of model, every hardware has its own software on-chip which provides all its required functionalities; and thus, is referred to as ad-hoc because it is made out of cohesive and tightly-coupled modules that are hard to be adapted for other purposes. On the other hand, a service-oriented architecture would decouple the software from the hardware and expose it in form of web service components through a server possibly located in battle control centers, operation rooms, or in space stations operated in low earth orbit. Military equipment, devices, and vehicles supporting computational combat operations can then remotely communicate with existing services to acquire their necessary functionalities. Additionally, using a service-oriented architecture, military equipment are no more monocoque systems composed of one single unit but of loosely-coupled distributed components that are separated from their physical hardware and hosted in a remote location.

## 3. SERVICE-ORIENTED ARCHITECTURE

Service-Oriented Architecture or SOA for short is a model for system development based on loosely-integrated suite of services that can be used within multiple business domains [13]. SOA is also an approach and practice for building IT software systems using interoperable services. These services are loosely-coupled software components that encapsulate functionalities and are available to be remotely accessed by client applications over a network or Internet [14]. The backbone of SOA consists of web services and an Enterprise Service Bus (ESB).

**3.1 Web Services**

As defined by W3C, a web service is a software component designed to support interoperable machine-to-machine interaction over a network [5]. It uses the SOAP, an XML-based protocol to communicate over HTTP. Characteristically, web services have three key elements: Web Service Description Language (WSDL) which is an XML-based description of the operations and functionalities offered by the web service. It dictates the protocol bindings and the message formats required to connect to and interact with a given web service; Universal Description, Discovery and Integration (UDDI) which is a registry for storing web services' WSDLs and a mechanism to register and locate web services on the Internet; and the SOAP communication protocol which defines the structure and format of the messages being exchanged between the service requester represented by the client and the service provider represented by the actual web service. In fact, the service requester is a client application requesting a particular functionality from the service provider, and the service provider is usually a server that hosts and runs the actual web service. Other types or styles exist for web services. They include REST, RPC, RMI, .NET Remoting, CORBA, and Network Socket [15].

REST (Representational State Transfer) web services do not use the SOAP protocol to communicate; rather, they use the plain HTTP protocol and Query String information to exchange messages. Their advantages over SOAP-based web services are that they are easier to build, manage, and reuse.

RPC (Remote Procedure Call) is an inter-process communication that allows a computer program to invoke or call remotely a function or procedure to execute on another computer over a shared network. RMI (Remote Method Invocation) is the Java implementation for RPC, while .NET Remoting is the .NET implementation for RPC.

Network Socket is an inter-process communication between two or more computer programs over a network. A server socket uses a socket address which is a combination of an IP address and a port number to listen for incoming connections. Clients connect to the server socket and then start exchanging data packets. Network sockets can be implemented using either TCP or UDP protocols.

Figure 1 illustrates the infrastructure of a generic web service.

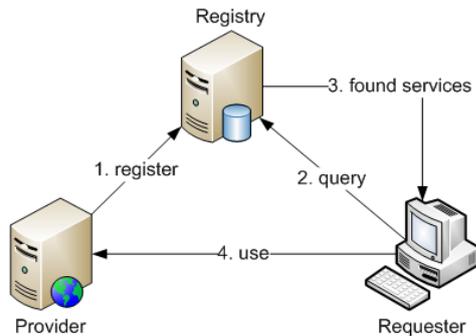

**Figure 1:** Infrastructure of a typical web service

### 3.2 Enterprise Service Bus - ESB

In order to promote interoperability among its components, SOA often employs an Enterprise Service Bus or ESB. Fundamentally, an ESB is a piece of software that lies between the different components of an SOA, mainly between the service requester and the service provider to enable a transparent and seamless communication among them [16]. It, in fact, acts as a middleware and a message broker between the different communicating parties in SOA architecture. The primary task of ESB is to support message routing and ensure a better orchestration and interoperability between the various interconnected web services possibly built using different technologies, platforms, standards, and programming languages. Figure 2 shows an ESB connecting incompatible consumers and producers built using different technologies.

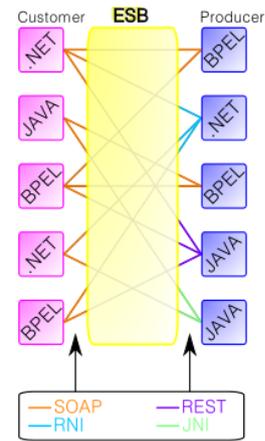

**Figure 2:** Architecture of an enterprise service bus

## 4. PROPOSED ARCHITECTURE

This paper proposes a Service-Oriented Architecture (SOA) for building weaponry and battle command and control systems using service software components. It is a distributed model made out of loosely-coupled interoperable web services and a central Enterprise Service Bus (ESB) not located inside the actual military hardware equipment but in an isolated location, possibly operation control centers or space stations in low earth orbit. The communication between the military equipment and the web services is bi-directional and is done in a remote fashion using the HTTP protocol with the help of the ESB acting as a middleware. The employed communication style is method invocation in which military equipment can remotely call or invoke the different procedures of the existing web services to execute on the hosting system and return results to the equipment. These procedures also known as methods or functions contain the logic and the programming instructions that deliver the basic functionalities for the military equipment. Essentially, the proposed architecture is composed of three basic tiers:

The first tier is the client represented by the military equipment or any weaponry system supporting computation, which invokes the different exposed methods of web services to perform a wide range of operations such as telemetry & tracking, ballistics calculations, launch control, aerospace traffic control, flight planning, surveillance and monitoring, fires and effects, logistics and mediation, intelligence and security, GPS and navigation, data acquisition, processing, and analysis, image processing, digital signal processing, data cryptography, and biometrics.

The second tier is the server represented by web services which are decoupled from the hardware of military equipment and hosted and executed on server machines located in battle control centers. The web services provide the actual code and logic for the different military operations and functionalities. They contain the algorithms, implementation, and programming instructions necessary to provide the various military computing machineries their basic maneuvers and functionalities.

The third tier is the middleware represented by the Enterprise Service Bus which offers a standard interface and a unified data-path for both the client and the server tiers to interoperate efficiently and exchange data regardless of their incompatible platforms and implementation technologies, for instance, technologies such as SOAP, REST, RPC or others. Figure 3 illustrates the proposed SOA architecture and its different tiers.

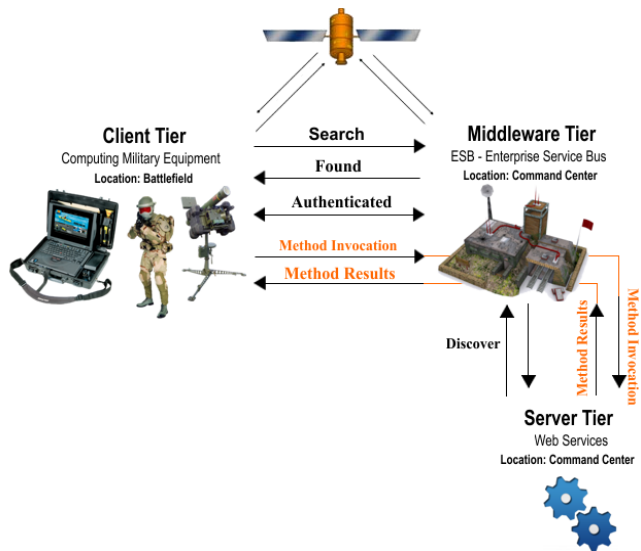

**Figure 3:** Different tiers of the proposed SOA architecture

### 4.1 Advantages & Motivations

A service-oriented architecture for battle command and control systems would decouple, isolate, and detach the software from the core hardware of military equipment, devices, and computing machines, making them independent and not physically bound to each other. As a result, military equipment are no more composed of a single block housing both hardware and software; rather, only hardware constitute the actual equipment; while, software consist of loosely-coupled distributed web services that encapsulate the basic military functionalities and are executed remotely outside the military equipment hardware. In other words, the military computing equipment only send requests to and get results from the various available web services. Basically, the proposed SOA design has many advantages which can be listed below:

Integrate-ability: Integration of new software components can take less significant effort, time, and budget. For instance, new services providing new functionalities can be easily deployed on the server tier without the need to access the out of reach military machines and weaponry equipment. Likewise, changes to the existing web services can be easily made by only changing the service description on the server side.

Scalability: SOA is an open architecture in that it supports plug-and-play operations. For instance, new services can be deployed at runtime with no or minimal amount of system interruption. Similarly, they can be pulled out of the system at any time without experiencing degradation in performance or shortcomings in system operation. On the other hand, existing services can be reconfigured and updated at minimal cost. As SOA is governed by the publish-discover process [17], delivering new services and consuming them is usually done in an automated manner.

Maintainability: Since services are no more part of the equipment hardware and thus located at a great distance away from the fighting sites, it is less tedious and less costly to isolate system defects and troubleshoot, diagnose, and repair broken services. Consequently, this promotes agile and robust systems that can cope with unpredictable and always changing environments without affecting the system in operation.

Reusability: Services can be reused to add or extend new functionalities or build new military systems from already existing components. This practice can reduce design, development, implementation, testing, and deployment time.

Decentralization: Being modular, SOA components can be dispersed over multiple hosting environments providing computing power over distributed and inexpensive machines of massive computing arrays.

Survivability: In warfare, military systems are always subject to numerous physical and electronic attacks. One key feature of SOA is the self-organizing provider-consumer peer-to-peer network model which allows web services to be replicated across and migrated between servers and deployed where they are needed at several sites. This ensures the continuous operation of the participating systems in spite of hostile attacks, hits, and bombings.

Interoperability: As SOA features an ESB which emulates a middleware that sits between the different military equipment and web services, it provides a standardized and cross-network platform over which computing military machines can interoperate transparently with numerous existing systems and with the different heterogeneous web services that are built using different standards, programming languages, technologies, and platforms.

### 4.2 The Client Tier – The Military Equipment

Actually, the client is any computing military equipment, device, machine, combat vehicle, aircraft, naval ship, communication system, infrastructure, computer, or smart phone used in the battlefield by both soldiers and commanders. They contain an onboard computer able to discover the different remote web services through the ESB interface which describes the different functions encapsulated within the connected web services. In order to communicate, the client equipment has to bind to the ESB interface. This binding authenticates the military equipment (requester) and allows it to send requests to the ESB (provider) using remote procedure invocation approach. All execution is done on the provider's side and only results are returned to the requester. Communication between requester and provider is done solely using the HTTP protocol through communications satellites that relay transmission between the earth where the provider is located and the battlefield where the requester is located. Figure 4 illustrates the sequence of interactions between the military equipment as client, the ESB as middleware, and the web services as server.

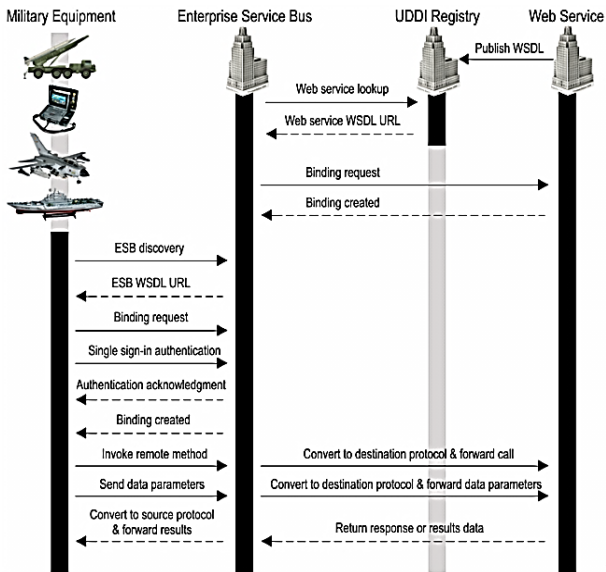

**Figure 4:** Sequence of interactions between the SOA entities

### 4.3 The Middleware Tier – The ESB

The ESB or Enterprise Service Bus provides a data-path for data to travel between the military equipment and the web services. It constitutes a data transmission medium, emulating a messaging middleware that links between the different military equipment in the battlefield from one side and the different distributed web services from the other side to allow them to send and receive data back and forth to each other. Additionally, it automates the in and out communications between all involved systems and coordinates the interaction between them, and allows the storage, routing, and transformation of messages during inter-system interactions. The proposed ESB is cross-platform and cross-network which allows the military equipment to interoperate with various types of web services, possibly incompatible and built using different platforms, different standards, different technologies, and different programming languages to send requests, and receive responses from each other. Figure 5 depicts the architecture of the proposed ESB together with its inner-workings.

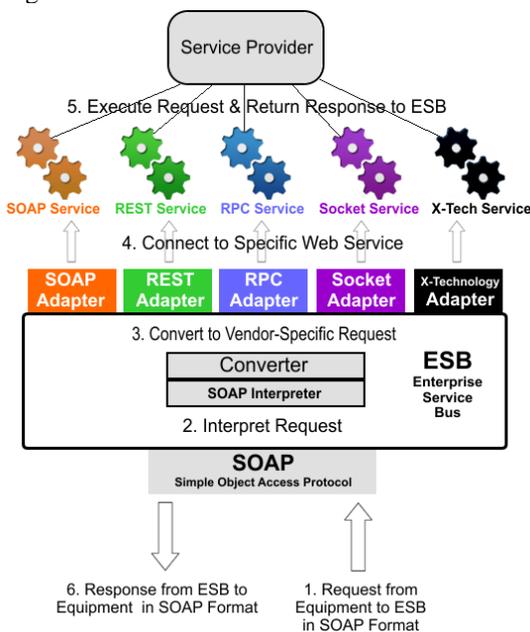

**Figure 5:** The architecture of the proposed ESB

In effect, the ESB has two public interfaces: The first interface is from the military equipment's side which provides a unified single SOAP-based end-point for the equipment to communicate with the ESB. The second interface is from the web services' side which provides a set of adapters as end-point connectors for the different web services to connect. There exists an adapter for every web service protocol, for instance, SOAP, REST, RPC, Network Socket, and others. The role of these adapters is to bridge the equipment's requests with their destination web services, irrespective of their protocol type and version. In order to achieve this, the ESB is able to identify the type of request received from the military equipment and to route it accordingly to the corresponding adapter which, successively, passes it to the corresponding web service. All in all, the ESB delivers a transparent communication between the different components of the SOA allowing them to interoperate despite their underlying incompatible technologies and platforms. The ESB communication process can be described as below:

Step 1: A military equipment invokes a function called *motion_detection()* located in a REST-based web service. The request is always in SOAP protocol and encapsulates metadata describing the request message, including the source client, the destination service, the function to call, and a set of parameters.

Step 2: The ESB receives the request message in SOAP format; it first validates the correctness of its XML structure and then converts it from SOAP format into the protocol of destination web service, in this case REST, using the protocol translator. The ESB uses an internal registry to lookup the technical details about the destination web service.

Step 3: The ESB routes the converted request to the adapter that is compatible with the addressed web service, in this case, the REST adapter.

Step 4: The adapter then locates the corresponding web service and gets bound temporary to it and starts executing the requested function, in this case *motion_detection()*.

Step 6: Once processing is done, a response is sent back from the destination web service to the military equipment that originally initiated the request. It is first sent to the corresponding adapter, in this case, the REST adapter, then to the ESB, then translated to a SOAP format, and eventually routed to the military equipment.

### 4.4 The Server Tier – The Web Services

The server tier is where the web services are hosted. It mainly consists of several mainframe computer servers often located in earth battle control centers. These servers define the execution of the web services, process military equipment's requests, execute business logic, and perform intensive calculations on behalf of the equipment. The web services can be of any type, protocol, or version and they interact with the ESB through its multi-platform end-point adapters. Each time a new web service is integrated into the system, it publishes its WSDL to the ESB which saves it inside an internal registry along with other important details. The ESB then exposes the WSDL to all military equipment allowing them to call remotely all available functions.

Web services can provide any type of functionalities including GSM to receive and transmit telemetry data between the different military units using SMS or other communication technologies; navigation to monitor and control the movement of combatting vehicles and determine their positions using radars, sensors, and satellites; ballistics computations to calculate the trajectories of projectiles, such as bullets, gravity bombs, rockets, or the like; imaging and computer vision to analyze captured images and recognize objects within these images, often useful for military reconnaissance and surveillance; and biometrics to authentic military units and provide identity access management and access control based on one or more inherent physical traits such as fingerprint, face recognition, iris recognition, and palm print [18].

## 5. EXPERIMENTS & IMPLEMENTATION

As a proof of concept, a client simulation software, representing a military equipment, was built and is capable of sending requests to and reading results from the ESB using the SOAP protocol. The software is a regular standalone executable application built using C#.NET under the .NET Framework 4.0 and the MS Visual Studio 2010. Figure 6 depicts the GUI interface of the client simulation software.

**Figure 6:** Simulated-client's GUI interface

Additionally, three web services were developed. The first is a SOAP-based web service built using C#.NET with an *.asmx* file extension and is capable of performing biometric operations. The second is a REST-based web service built using Java with a file extension *.jsp* and is capable of performing ballistics computations. The third is a Socket-based web service built using C++ with an *.exe* file extension and is capable of performing GPS operations. Figure 7 is a source-code snippet for a method extracted from the SOAP-based web service whose aim is to convert a scanned fingerprint into a bitmap image so that it can be digitally processed.

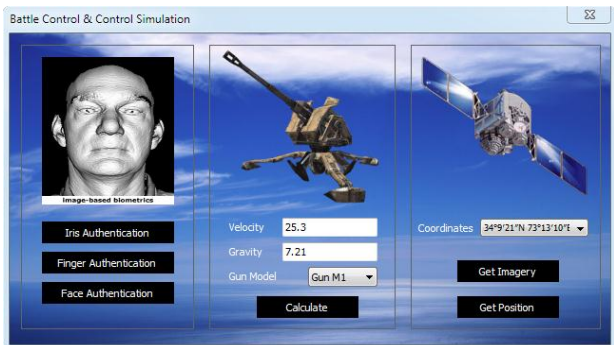

**Figure 7:** Fingerprint processing method

Finally, an ESB was built to act as a middleware between the simulated client and the different web services. Since the ESB acts as a service broker, it is responsible for exposing the various web services functionalities to the simulated client. Figure 8 delineates the list of functionalities exposed by the ESB and originally implemented in the web services.

**Figure 8:** Various methods exposed by the ESB

For verification purposes, a use case scenario [19] was created. Its purpose is to test the validity and the interoperability of the client-web-service communication through the ESB.

1. The client simulation software needed to execute a function called *Compute_Trajectory()* located in the REST-based web service, so it connected to the ESB in a process to discover all public available functionalities.
2. Once function *Compute_Trajectory()* was found, the client bound to the ESB and sent an authentication message to the ESB.
3. The ESB acknowledged the client allowing it to start remote function invocation.
4. The client invoked function *Compute_Trajectory()* sending *gravity=9.8* and *velocity=45* as parameters to the ESB using the SOAP protocol.
5. The ESB received the call and then looked-up for the destination web service that encapsulates function *Compute_Trajectory()*.
6. Once the corresponding REST-based web service was found, the ESB converted the client's call message from SOAP into REST and forwarded it to the web service.
7. The REST-based web service received the call, directly processed it, and executed function *Compute_Trajectory()* on its hosting server.
8. Upon finishing processing, the web service returned the result *angle=14.12* to the ESB in REST format.
9. The ESB converted the REST message into a SOAP message that is readable by the client, and forwarded it to the client.
10. The client received the result and displayed it on the screen.

Furthermore, other use cases were executed at runtime while the system was running, and in all situations the client succeeded to adapt itself according to the new changes in the environment. The different uses case scenarios that were tested are given below:

1. Integrating a new web service
2. Removing an existing web service
3. Updating web service functionalities
4. Failing an existing web service
5. Fixing a faulty web service
6. Deriving new web services from existing ones

## 6. VALIDATION OF THE PROPOSED ARCHITECTURE

The SOA approach proved to be very effective in all the different executed scenarios. The interoperability of the system allows the collaboration between various entities regardless of their underlying technologies and implementation details. The scalability of the system allows the military specialists to easily and quickly alter and add functionalities to military equipment without having access to them. The maintainability of the system allows fixing and replacing out of order services while the system is running with no or minimal operation interruption. The reusability of the system allows building and deriving new web services from existing ones with the least amount of development time and cost.

## 7. CONCLUSIONS & FUTURE WORK

This paper presented a novel service-oriented architecture for building battle command and control systems using distributed software components called web services. The proposed architecture consists of three tiers: the client tier corresponding to any sort of computing military equipment that require executing some functionalities; the server tier corresponding to the web services that deliver the basic functionalities and operations for the military equipment; and the ESB acting as a middleware that coordinates and shields the complexity and heterogeneity of communication among the different entities of the system. Experiments conducted showed a robust, reliable, scalable, interoperable, reusable, and a maintainable architecture that can adapt itself to the unforeseen circumstances and cope with the various obstacles that might be encountered during war fighting missions.

As future work, an encryption layer is to be added to the proposed SOA architecture so as to protect and conceal the exchange of messages and data communication between the various entities of the system.

## ACKNOWLEDGMENTS

This research was funded by the Lebanese Association for Computational Sciences (LACSC), Beirut, Lebanon, under the "Service-Oriented Architecture Robotics Research Project – SOARRP2012".